\documentclass[runningheads]{llncs}
\usepackage[misc]{ifsym}
\usepackage{hyperref}
\usepackage{svg}
\usepackage{amsmath}
\usepackage{amssymb}					
\usepackage{placeins}
\usepackage{graphics}
\usepackage{graphicx}
\usepackage{times}
\usepackage[utf8x]{inputenc} 

\usepackage{xspace} 
\newcommand{\CYCA}{\texttt{CYCA}\@\xspace}
\newcommand{\CYCAL}{\texttt{CYCA-L}\@\xspace}
\newcommand{\CYCAS}{\texttt{CYCA-S}\@\xspace}
\newcommand{\RLYCA}{\texttt{RLYCA-A}\@\xspace}

\begin{document}
\title{An Improved Yaw Control Algorithm for Wind Turbines via Reinforcement Learning}
\titlerunning{An Improved Yaw Control Algorithm for Wind Turbines via Reinforcement Learning}


\author{Alban Puech \inst{1,2} \Letter \and Jesse Read\inst{1} }


\authorrunning{Alban Puech \and Jesse Read}

\institute{
LIX, Ecole Polytechnique, Institut Polytechnique de Paris \\
Palaiseau, France \\
\and DEIF Wind Power Technology Austria GmbH \\
Klagenfurt, Austria \\ }

\toctitle{An Improved Yaw Control Algorithm for Wind Turbines via Reinforcement Learning}

\maketitle

\begin{abstract}
Yaw misalignment, measured as the difference between the wind direction and the nacelle position of a wind turbine, has consequences on the power output, the safety and the lifetime of the turbine and its wind park as a whole. We use reinforcement learning 
to develop a yaw control agent to minimise yaw misalignment and optimally reallocate yaw resources, prioritising high-speed segments, while keeping yaw usage low. To achieve this, we carefully crafted and tested the reward metric to trade-off yaw usage versus yaw alignment (as proportional to power production), 
and created a novel simulator (environment) based on real-world wind logs obtained from a REpower MM82 2MW turbine. The resulting algorithm decreased the yaw misalignment by 5.5\% and 11.2\% on 
two simulations of 2.7 hours each, compared to the conventional active 
yaw control algorithm. The average net energy gain obtained was 0.31\% and 0.33\% respectively, compared to the traditional yaw control algorithm.  On a single 2MW turbine, this amounts to a 1.5k-2.5k euros annual gain, which sums up to very significant profits over an entire wind park.

\end{abstract}

\section{Introduction}
As the world tries to move away from fossil fuels, wind energy appears to be one of the most promising renewable energy sources. Wind turbines convert kinetic energy from the wind into electricity. Energy output mainly depends on the wind speed, the turbine blade diameter and the generator size, but it is now known that the layout, as well as the operation of turbines, can have a significant impact on the power performance of wind parks \cite{sig_wt,effect_ye}. In particular, the orientation of the wind turbine rotor against the wind, achieved by the yaw mechanism, is of key importance to ensure that the maximal amount of wind energy is extracted. To do so, one needs to minimise the yaw misalignment $\gamma$, or yaw error, measured as the difference between the wind direction $\phi$ and the nacelle position $\theta$:
\begin{equation}
\label{yaw_mis}
\gamma = \phi - \theta
\end{equation}
Based on a collection of 300 turbines, \cite{per_ym} showed that more than 50 \% of wind turbines were misaligned to the wind direction, while in \cite{cost_ym}, the loss yield by a misalignment of only 7 degrees on a 2MW wind turbine was estimated to be around 6000 euros/year.
Yaw misalignment also has an impact on the wind park as a whole since it increases the wake effect left by the rotor on the downstream turbines, as shown by \cite{effect_ye_load}.
Finally, yaw misalignment reduces the lifespan of the turbine, putting them under high constraint loads \cite{cube3}.

The yaw control algorithm that we use in the present paper as a baseline is a conventional yaw control algorithm (\label{CYCA}\texttt{CYCA}) deployed in most modern fixed and variable speed horizontal axis wind turbines. In this paper, the results of \CYCA are obtained in two different ways. When needed, we make the distinction between the version of \CYCA running on the wind turbine and which control sequence (sequence of control action) is constructed from the turbine logs (\label{CYCAL}\CYCAL), and its simulated version \label{CYCAS}\texttt{CYCA-S}, which control sequence is obtained using a yaw control simulator from DEIF. It is important to note that both \CYCA versions correspond to the same algorithm, but their results differ due to external factors (e.g. maintenance, cable untwisting, emergency stop) that interfere with the correct execution of \CYCA while running on the turbine. \CYCA takes the wind direction measured over a small duration preceding the yaw control as an estimate for the nacelle position to reach during the next control cycle. It typically suffers from irreducible operation delays stemming from lags in communications and its fixed (relatively slow) yaw speed. Beyond slow reaction time to wind direction change, the yaw resources are intentionally limited by the controller because a too-frequent yaw usage risks damage to key components of the mechanism such as the bearings. 

\begin{figure}[ht]
\centering
\includegraphics[width=\textwidth]{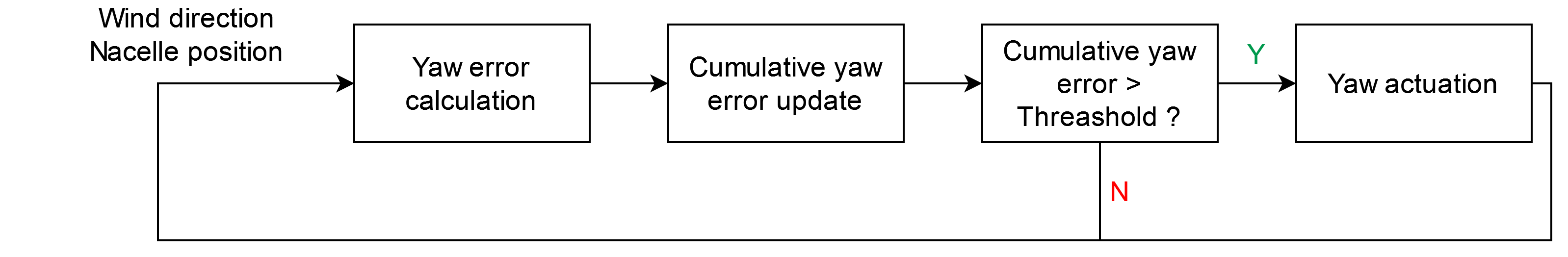}
\caption{Simplified control loop of \CYCA. The cumulative yaw error is updated with the new yaw error computed from the wind direction measurement. The value of the threshold defines the sensibility of \CYCA to small or short yaw misalignment. The length of the control cycle varies but is usually around 1 second.} 
\label{trad_diagram}
\end{figure}

To limit unnecessary yaw actuation and load on the mechanism, \CYCA actuates the yaw mechanism only after the cumulative error of yaw misalignment exceeds some threshold. As a result, the same importance is given to yaw alignment in low-speed segments, where the potential energy gain by reducing yaw misalignment is low, and to higher speed segments, where most of the energy gain can be achieved. We summarise the working principle of \CYCA in Fig.~\ref{trad_diagram}. An optimal yaw system control algorithm would allocate yaw resources in order to minimise the yaw misalignment, with a priority given to the regions where most of the energy gain can be obtained.

Little research has been done on new yaw control strategies compared to other turbine operations, such as pitch control. Some yaw control strategies rely on direct or indirect measurement of the wind direction, while others rely on maximum power point tracking. Most of them depend on wind measurements that are sensitive to external factors. Moreover, yaw alignment is often achieved at the expense of yaw usage. They also do not consider the power output for the allocation of the yaw resources. To address those issues, we propose in this paper a new data-driven yaw control strategy that bases the control decision on prediction and pattern recognition rather than past measurements. This new strategy, relying on a Reinforcement-Learning (RL)-trained agent, increases the power output by reducing the overall yaw misalignment and by intelligently re-allocating yaw resources. The contributions of this paper are:
\begin{itemize}
\item A novel RL-based yaw control algorithm for wind turbines that optimally allocates yaw resources and better tracks wind direction change, keeping yaw usage low.
\item A new realistic simulation yaw environment that allows training and benchmarking of our algorithm against several baselines (simulated and real-world conventional yaw control algorithm) using past wind conditions obtained from turbine logs.
\item A new multi-objective reward function to handle short wind direction variations without the use of deadband functions and new evaluation metrics that take into account the yaw drive consumption changes.
\end{itemize}
We organise the remainder of the paper as follows: We will first introduce the novel Reinforcement-Learning control algorithm \label{RLYCA}(RLYCA) and discuss its key components. We will then focus on the RL environment used for both training and assessment of the novel strategy. Finally, we will discuss the results of the proposed strategy and the possible improvement opportunities.


\section{Related Work and Existing yaw control strategies }
There are mainly 4 different types of active yaw control strategies. Contrary to passive yaw control strategies, which are based on wind force to orient the rotor, active yaw control strategies use a servomechanism to change the rotor position. Among the active yaw strategies, three of them are deterministic: the ones with advanced measurements, without wind direction measurement and with indirect measurements. Recently, RL methods were proposed and output a control action based on pattern recognition. \\

{\bfseries Yaw control with advanced measurements}. Yaw control with advanced measurements relies on remote instruments like Lidars and Sodars to get more accurate wind direction and speed values, as in \cite{cost_ym}. As the wind direction measure is made upstream of the turbine, it is not affected by the wake effect and gives a more accurate measure of the yaw misalignment, which is then used by an active yaw control algorithm. Alternatively, a nacelle-mounted Lidar can be used to learn a correction function that is then applied to the wind direction measurement obtained from the wind vane \cite{lidar}. Both methods were shown to help at correcting the yaw angle, but they require expensive measurement tools.\\

{\bfseries Yaw control without wind direction measurement}.
Other methods, without wind direction measurement, have been proposed recently \cite{without_w,without_w2}. They rely on maximum power point tracking. The difference between the actual power output and the theoretical power output computed using the power curve gives information on the power loss coming from the yaw misalignment, and, in turn, gives an estimate of the yaw misalignment. A benefit of those methods is that they no longer depend on the wind direction measurements that are often, as discussed previously, disturbed by the operation of the wind turbine and by the wake effect. However, they are sensitive to wind speed variation since the theoretical output power is computed out of it and to any other external factor such as snow or dirt that could reduce the power output of the turbine. \\

{\bfseries Yaw control with indirect measurement}. The methods that gained the most attention recently are the ones based on indirect measurements. They rely on short term prediction of the wind direction to feed a model predictive control (MPC) algorithm that anticipates wind direction change \cite{mpc1,mpc2,mpc3}. Although the algorithm is no longer directly using the wind direction measure, its performance is still highly dependent on the accuracy of the wind direction prediction. In high wind variability conditions, the accuracy tends to be low (Dzulfikri, Nuryanti and Erdani \cite{mpc1} showed a mean absolute error of 9.9 degrees for a 1-minute ahead prediction of the wind direction on the testing set), which restricts the use of such methods to a relatively small space of wind conditions. It is important to note that, to this day, none of the proposed methods relying on wind direction prediction significantly outperforms the traditional active yaw control strategy at reducing the yaw misalignment. \\

{\bfseries RL-based yaw control strategies}.
Finally, the RL methods are an alternative to the other data-driven methods presented previously as they do not use the wind direction and speed measurements in a predefined, deterministic way. They rather determine the best control action based on the current wind and position state, learning from different simulation scenarios or online, while running on the turbine controller. Those techniques are the most promising as they always output the optimal control action to the best of the model knowledge acquired from the training. They work in a black box manner and are not impacted by the measurement inaccuracy since no control strategy needs to be designed in advance, and since no decision relies directly on the value of the wind direction or speed. The performance of the models then depends on the abstraction and generalisation potential of the algorithm and the quantity and relevance of the training data for the given task, which can be a limitation in situations where we do not have enough data. Those strategies make easier the task of reallocating yaw resources by prioritising segments of high potential gain since the algorithm can learn to identify those segments.

To this day, most of the research proposing an RL-trained agent focuses on long term control. Saenz-Aguirre \cite{rl1} proposed a model that actuates the nacelle position every 24 hours only. This present paper shows a method that is suitable for most wind turbines that usually spend 3 to 10 \% of the time yawing. \\
In addition to that, the wind conditions used in the simulators often make the control decision easier to determine. The wind direction used in the simulations ran to test the control algorithms against \CYCA often has the same mean over any segment of the dataset and shows periodic sudden change with some added noises of small magnitude. The wind speed is often taken to be almost stable with very small variations around its mean \cite{rl1,rl2}. Instead, we base our simulations on real-world data obtained from turbine logs. Most of the papers proposing new yaw control techniques do not show their performances for different levels of yaw usage and the bench-markings against current yaw control strategies are often limited. In the present paper, we offer a clear comparison with both \CYCAS and \CYCAL and we take into account the power consumption change of the yaw drive in the computation of the energy gains.\\
The originality of this paper is that it offers a yaw control algorithm that can directly and realistically replace \CYCA, increasing the power extracted while keeping the yaw usage reasonable.


\section{Proposed Yaw Control Strategy}
\subsection{Goals and constraints of the strategy}
\begin{figure}[ht]
\centering
\includegraphics[width=0.5\textwidth]{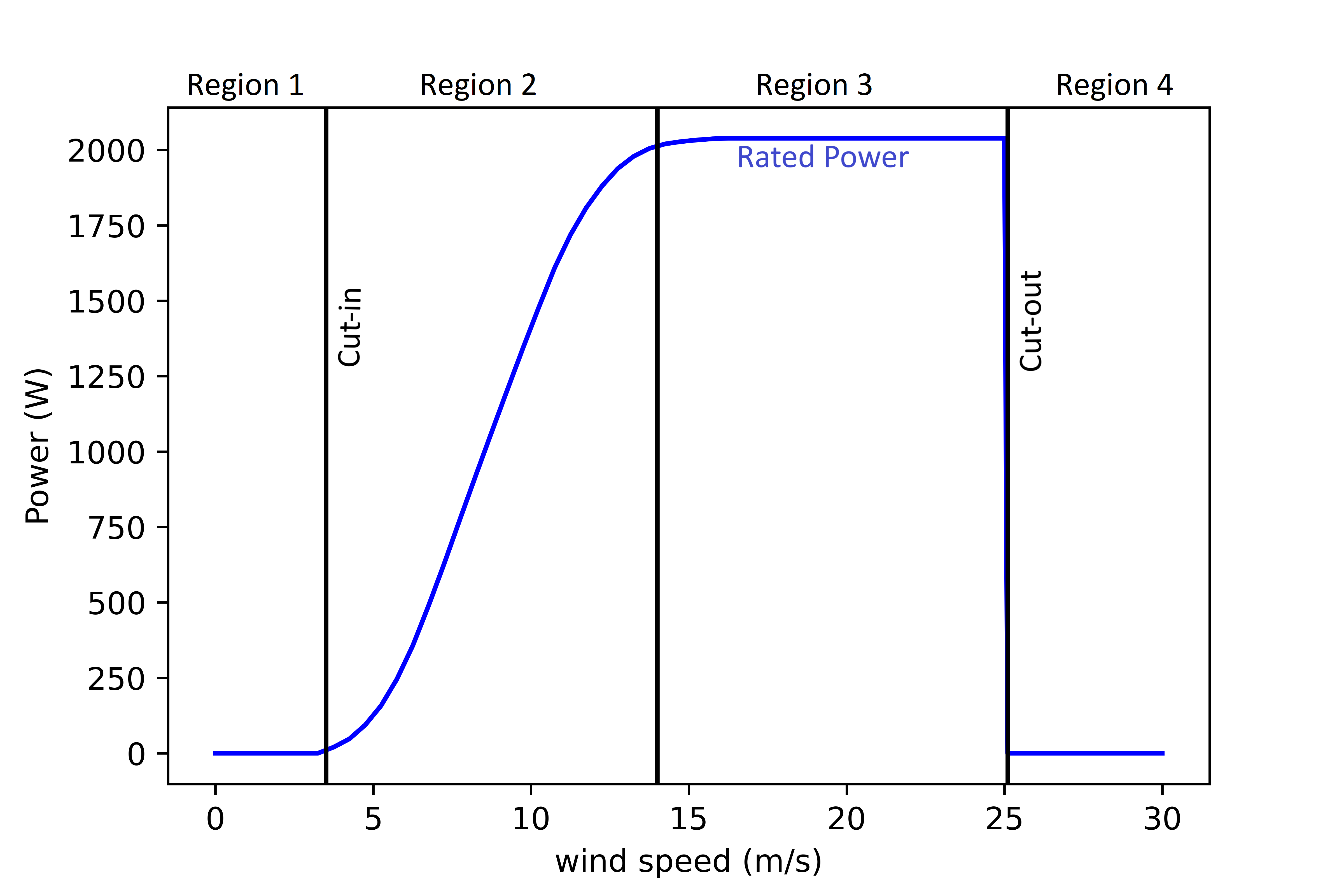}
\includegraphics[width=0.35\textwidth]{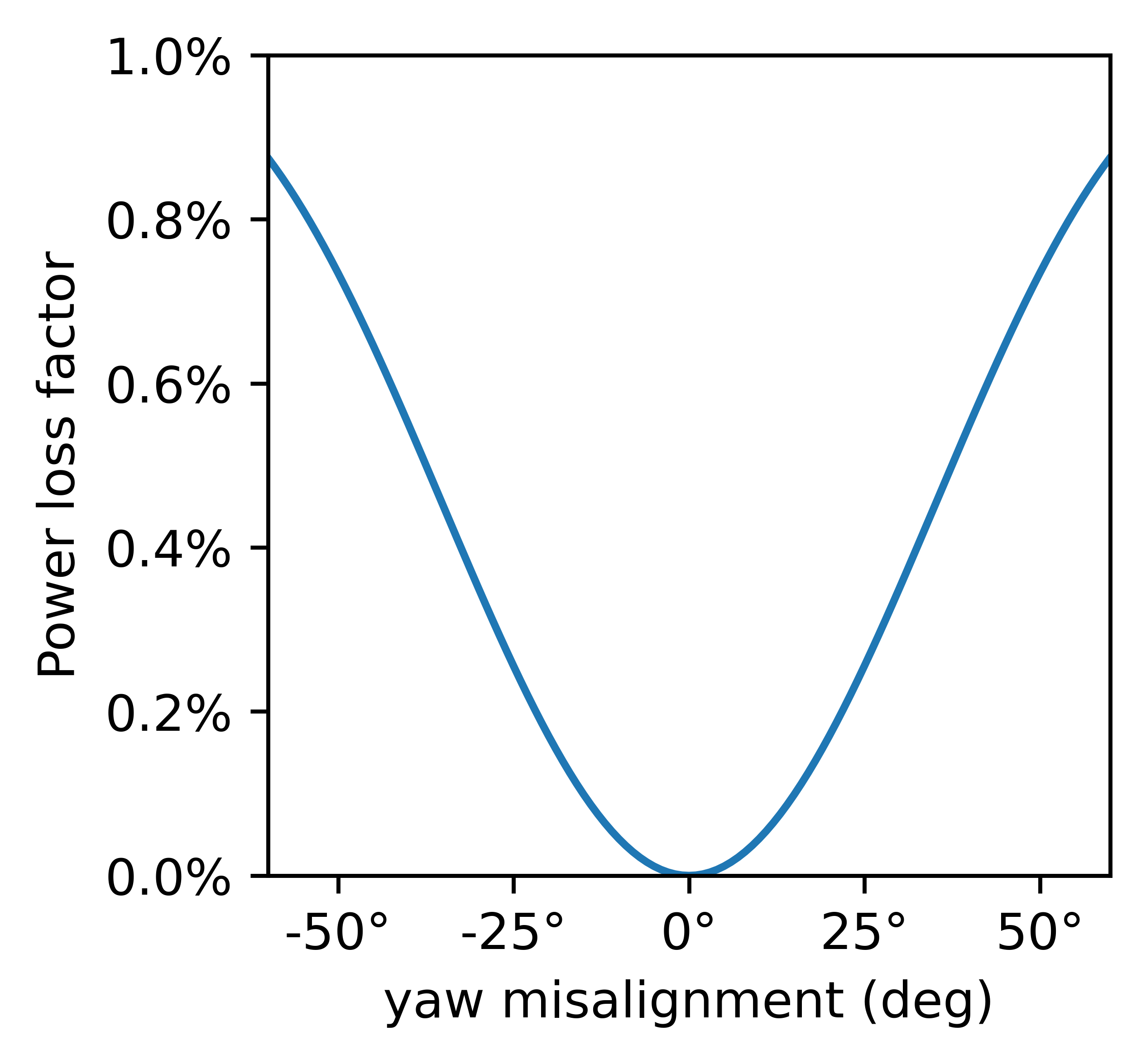}
\caption{Left: Power curve of 2MW turbine and its different regions. The turbine operates only after cut-in and before cut-out. The rated power corresponds to the maximum power the turbine can produce. Right: Power loss factor as a function of the yaw misalignment with $\alpha = 3$.}
\label{power_curve}
\end{figure}
One of the main motivations for our new yaw control algorithm is to maximise the wind power extracted by the turbine. Power loss due to yaw misalignment occurs in region 2 of the power curve, shown in Fig.\ref{power_curve}. Indeed, in region 1, that is, before cut-in, the wind is too slow and the turbine is not generating electricity. In region 4, after cut-out, the turbine is shut down and the rotor is braked to avoid over-speeding. Moreover, yaw misalignment does not have a significant impact on the power production in region 3, since the turbine is already outputting the maximum amount of power that it can output. However, the turbine will require a larger wind speed to reach rated power in case of yaw misalignment. In region 2, the power $P_\gamma$ generated by a wind turbine with a yaw misalignment $\gamma$ is expressed in degrees as:
\begin{equation}
\label{loss}
P_\gamma = P_* \cos^\alpha\gamma
\end{equation}
where $P_* $ is the power extracted assuming perfect yaw alignment. Moreover, $P_*$ is directly related to the wind speed $v$, the circular swept area of the turbine $A$, the air density $\rho$ and the Betz factor \cite{book} $c$: 
\begin{equation}
\label{power}
P_* = \frac{1}{2}\rho A v^3 c
\end{equation}
Many values have been proposed for the cosine exponent $\alpha$. Its value significantly impacts the calculated loss caused by yaw misalignment and it is often assumed that the value of $\alpha$ is three, as supported by \cite{cube1} and by analysis conducted by \cite{cube2}. Most recent studies like \cite{cube3} showed that the wake effect caused by upstream turbines can have an impact on the value of $\alpha$. Dahlberg and Montgomerie \cite{range} showed that the value of the cosine exponent could range from 1.7 to 5.1. Thus, we will stick $\alpha=3$, corresponding to the theoretical analysis proposed in \cite{cube2}. The power loss factor associated with a yaw misalignment in region 2 is shown in Fig.~\ref{power_curve}. The other benefit of reducing the yaw misalignment is to prevent the turbine from undergoing large mechanical and fatigue loads. Damiani \cite{storm} showed that misaligned turbines are more subject to damage in case of a storm but also see their lifespan reduced. Fleming \cite{load_increase} showed an increase in the blade, yaw bearings and tower loading in a situation of yaw misalignment.

Yaw misalignment minimisation should not be achieved at the expense of yaw usage. Indeed, yawing too often and for too long can harm the yaw mechanism. It is thus crucial to keep the time spent yawing under some limit mainly defined by the turbine safety guidelines, in our case, less than 10 \% of the time should be spent yawing. Hence, the goal is not to achieve perfect yaw alignment, but rather to find the optimal trade-off between yaw correction and yaw actuation. The main constraint is to avoid actuating the yaw systems too often and for too long. As the loss induced by yaw misalignment is cubic, reducing the yaw misalignment when it is large already allows us to decrease most of the loss caused by the misalignment. Moreover, and as discussed earlier, most of the yaw usage should be done on high-speed segments, where yaw misalignment has a higher cost on the power output.

\subsection{A proposed yaw control strategy via Reinforcement Learning}
Reinforcement learning (RL) involves teaching an agent progressing in an environment to take actions that maximise expected gains defined by a reward function. The agent has limited information concerning its current state and learns its policy based on experience from repeated episodes.

\begin{figure}[ht]
\centering
\includegraphics[width=200pt]{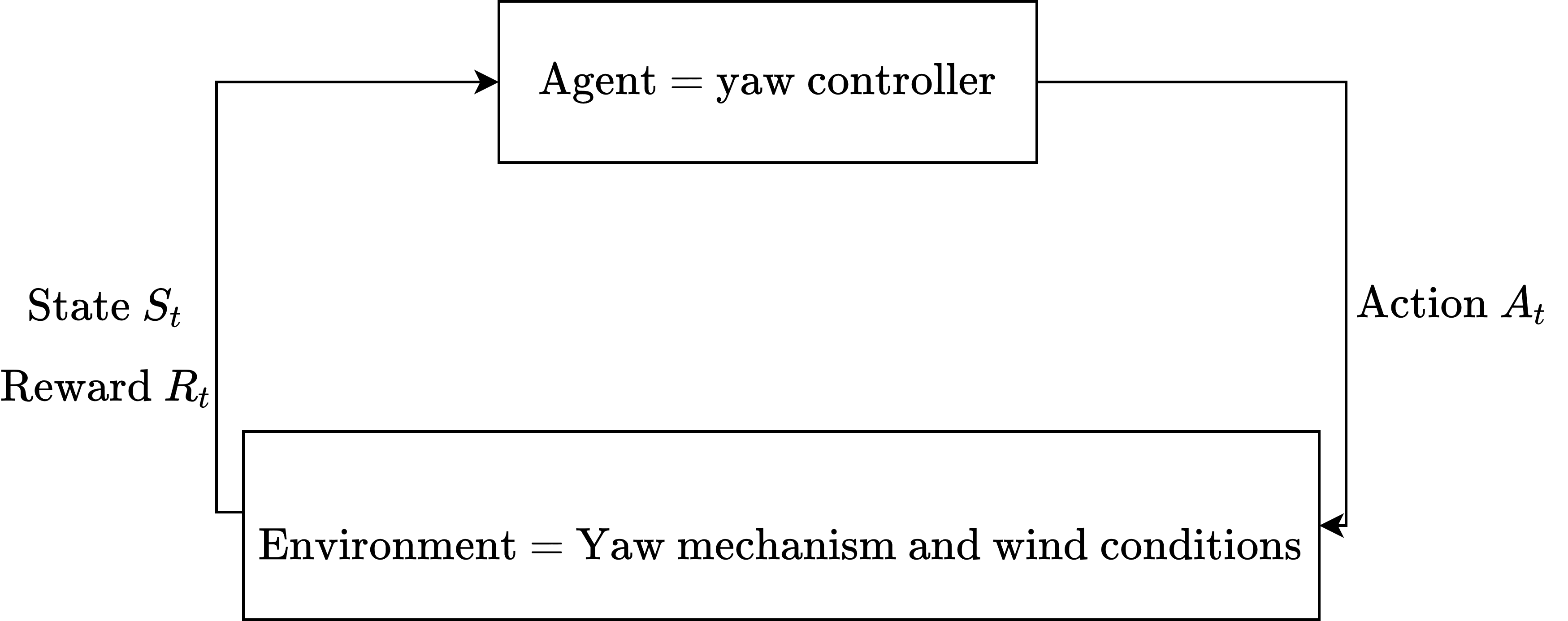}
\caption{Simplified block Diagram of the RL task. The new state $S_{t+1}$ and the reward $R_{t+1}$ is computed only after the agent took its action $A_t$.}
\label{rl_diagram}
\end{figure}

We define a control cycle as the episode of fixed duration (e.g. 3 seconds) during which information is received by the agent which then outputs a control action.
In the case of \RLYCA, during a control cycle, the agent takes an action $A_t$ by applying its internal policy, upon observing its current state $S_t$. After the action is taken, a reward $R_{t+1}$ and a new state $S_{t+1}$ is computed. The corresponding RL task can be described by Fig.~\ref{rl_diagram}.
We train the agent to learn the optimal control policy for the yaw system. Optimal control theory and reinforcement learning are closely related as they both refer to techniques used to find a control for a system that evolves in time and interacts with its environment, such that a certain objective function is maximised. However, an optimal control algorithm takes an action that maximises the predefined objective function in a deterministic way. In reinforcement learning, the agent does not know the reward function in advance and has to learn it through experiences to define its policy.

In this paper, the proposed yaw control strategy is based on a Proximal Policy Optimisation Algorithm (PPO) \cite{ppo}. Unlike Deep Q-learning, used in \cite{rl1,rl2}, PPO can not learn from offline stored data and requires online training. \\

\begin{figure}[ht]
\centering
\includegraphics[width=\textwidth]{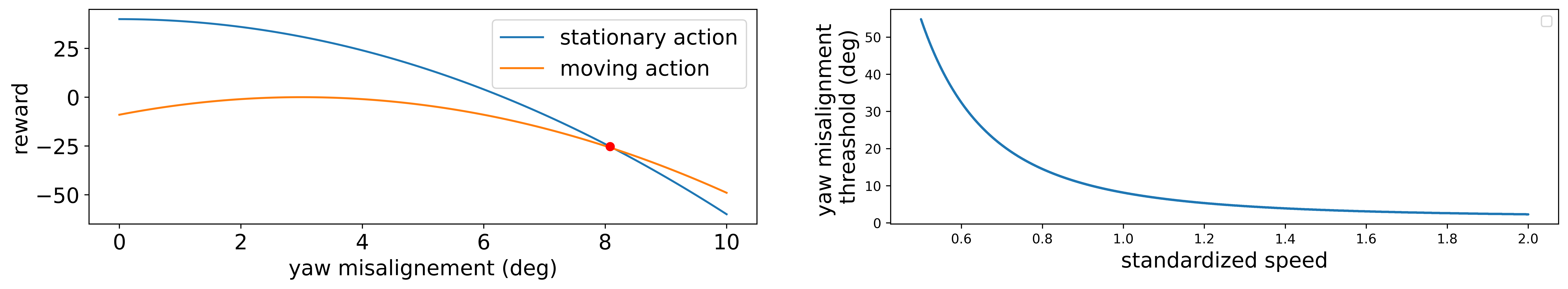}
	\caption{Left: Reward for the stationary action and the moving action as a function of the yaw misalignment under fixed $\tilde{v}=1$ and $w=40$. Right: Minimum yaw misalignment (vertical axis) starting from which the moving action has a higher reward than the stationary action (for $w=40$); as the wind speed increases, we encourage more and more yaw correction.}
\label{reward_func}
\end{figure}

{\bfseries Designing the action space.}
At each control cycle, the agent outputs one of the following actions:
\begin{itemize}
\item 0: Clockwise rotation for the full duration of the control cycle
\item 1: No rotation
\item 2: Counterclockwise rotation for the full duration of the control cycle
\end{itemize}
{\bfseries Defining the reward function.} The goal of the agent is to learn the optimal policy, which is the one that maximises the expected return, computed as the sum of rewards over an episode. We define a reward function in two parts, each accounting for one of our two goals: maximise power production, and limit the yaw mechanism usage. \\
The first term $R^{(1)}$ of the reward function accounts for the power production maximisation aspect. To maximise the power extracted during the control cycle $t$, we need to minimise the yaw misalignment $\gamma_t$ as expressed in Eq.~\eqref{yaw_mis}. However, more importance should be given to the control cycles where the wind speed $v_t$ is high (but still lower than the rated speed), as Eq.~\eqref{power} shows that the power extracted assuming perfect alignment $P_{0}$ increases with the cube of the wind speed $v$. Hence, we define $R^{(1)}_{t+1}$, the first negative component of the reward for the action taken during the control cycle $t$:
\begin{equation}
\label{rew1}
R^{(1)}_{t+1} = - \gamma_{t+1}^2 \tilde{v}_{t+1}^3
\end{equation}
Weighting the yaw misalignment $\gamma_{t+1}$ squared by the standardised wind speed $\tilde{v}_{t+1}$ cubed allows us to penalise more yaw misalignment when the power output is high, and thus encourages a better yaw resource allocation\\
The second term of the reward function, denoted by $R^{(2)}$, accounts for the second objective of \RLYCA: yaw usage limitation. To keep the yaw count low and reduce the time spent yawing, we should prevent the agent from reacting to short wind direction variations (in time) or to small yaw misalignments (in magnitude) that are not worth being corrected compared to larger yaw misalignment. Indeed, since the loss increases with the cube of the yaw misalignment, a correction of one degree will yield more gain when the yaw misalignment is large than when it is small. We thus want our agent to react in priority to large yaw misalignment, and to spare the yaw mechanism the rest of the time. Hence, we reward the agent for observing a period of at least $k$ control cycles without taking any moving actions. We define $R^{(2)}_{t+1}$, the second component of the reward for the action $a_t$ taken during control cycle $t$ as 
\begin{equation}
\label{rew2}
R^{(2)}_{t+1} = w \prod_{i=t-k+1}^t \mathbf{1}\{ a_i = 1\}
\end{equation}
where the weight $w \in \mathbb{R^+}$ encodes the importance given to the preservation of the yaw mechanism.To illustrate the principle of this loss function, let us assume that $w=40$, that the controller has not taken a moving action in the last $k-1$ control cycles, and that the wind conditions remain similar during the control cycle. Assume further that the maximum angle covered by the yaw mechanism in one control cycle is 3 degrees. If the agent takes the correct moving action (yaws in the correct direction), the yaw error can be corrected by a maximum of 3 degrees. The second component of the reward function will be 0, and the reward will be given by $ R_{t+1}(moving) = - (\gamma_{t}-3)^2 \tilde{v}_{t}^3$.\\
If the agent chooses the stationary action, the reward will be given by the sum of the two reward components defined in Eqs.~\eqref{rew1} and \eqref{rew2}, where the yaw misalignment remains the same since no correction was made and since we assumed stable wind conditions. The reward will thus be: $ R_{t+1}(stationary) = - (\gamma_{t})^2 \tilde{v}_{t}^3 + 40$.\\
Fig.~\ref{reward_func}~(left) shows the reward if a stationary action is taken (blue), and if a moving action is taken (orange). For the moving action to give a larger reward than the stationary action, the yaw error needs to be larger than 8.1 degrees, where the two curves intersect. This can be interpreted as the fact that the agent will decide to move the yaw mechanism only if the yaw misalignment is larger than this value, and will not actuate the yaw mechanism otherwise. Fig.~\ref{reward_func}~(right) shows, for different standardised wind speed values, the minimum yaw misalignment for the moving action to give a larger reward than the stationary action. We designed the reward function such that, when the wind speed is high, the agent is more willing to correct small yaw misalignment. Note that in region 3 of the power curve, the reward function penalises yaw misalignment in the same manner as described before, but not for the same reason. Indeed, although there is no power loss due to yaw misalignment in this region, the constraint loads are higher, which justifies yaw misalignment to be equally penalised.

{\bfseries Choosing the state space.} The reward being dependent on the yaw misalignment, the wind speed and the past actions, we aim at including that information into the state $S_t$ given to the agent to take its action $A_t$ during the control cycle $t$. We define $j$ as the number of lagged values or ancestor values given for each feature of the state. $S_t$ is thus expressed as in Eq.~\eqref{state}, where $\gamma_t$, $\phi_t$ and $\tilde{v}_t$ denote the yaw misalignment, the wind direction and the standardised wind speed during the control cycle $t$, respectively.
\begin{equation}
\label{state}
S_t = \begin{bmatrix}
a_{t-1} & \gamma_t & \phi_t & \tilde{v}_t \\
a_{t-2} & \gamma_{t-1} & \phi_{t-1} & \tilde{v}_{t-1} \\
a_{t-3} & \gamma_{t-2} & \phi_{t-2} & \tilde{v}_{t-2} \\
\vdots & \vdots & \vdots & \vdots \\
a_{t-j} & \gamma_{t-j+1} & \phi_{t-j+1} & \tilde{v}_{t-j+1}
\end{bmatrix}
\end{equation}
It is important to note that, in the control cycle $t$, 
state $S_t$ 
contains the wind direction $\phi_t$ and wind speed $v_t$, but 
not $v_{t+1}$ nor $\phi_{t+1}$ 
(the agent cannot know in advance the future wind direction and speed). 

\section{Yaw system simulation environment}
We develop a yaw system environment that we use with both \RLYCA and \CYCAS. In the case of \RLYCA, we use the environment for both training and testing. In the case of \CYCAS, we use the yaw system environment for testing only as the latter algorithm does not require to be trained. Finally, \CYCAL testing does not require any simulator since the control sequence, as well as all the measurements needed for the bench-marking, are directly extracted from the turbine logs.
\subsection{Datasets}
The yaw system simulation environment simulates the wind conditions recorded at a real-life wind park. 
Namely, we collected two time-series data sets each consisting of 21000 turbine log points, recorded in September and February of 2021. 
Each $t$-th data point ($t=1,\ldots,21000$) contains the mean wind direction and the mean wind speed over 1 second, stacked into state $S_t$ as illustrated in \ref{state}.  
Dataset 1 exhibits a relatively steady wind direction; of $34.1^\circ \pm 9.7$ with an interquartile range of $13.7^\circ$. Dataset 2 has similar statistical features (wind direction of $41.4^\circ \pm 11.6$ with an interquartile range of $14.2^\circ$) but shows fast wind direction changes of large magnitude between 10000s and 12500s and between 15000s and 20000s.

\begin{figure}[ht]
\centering
\includegraphics[width=\textwidth]{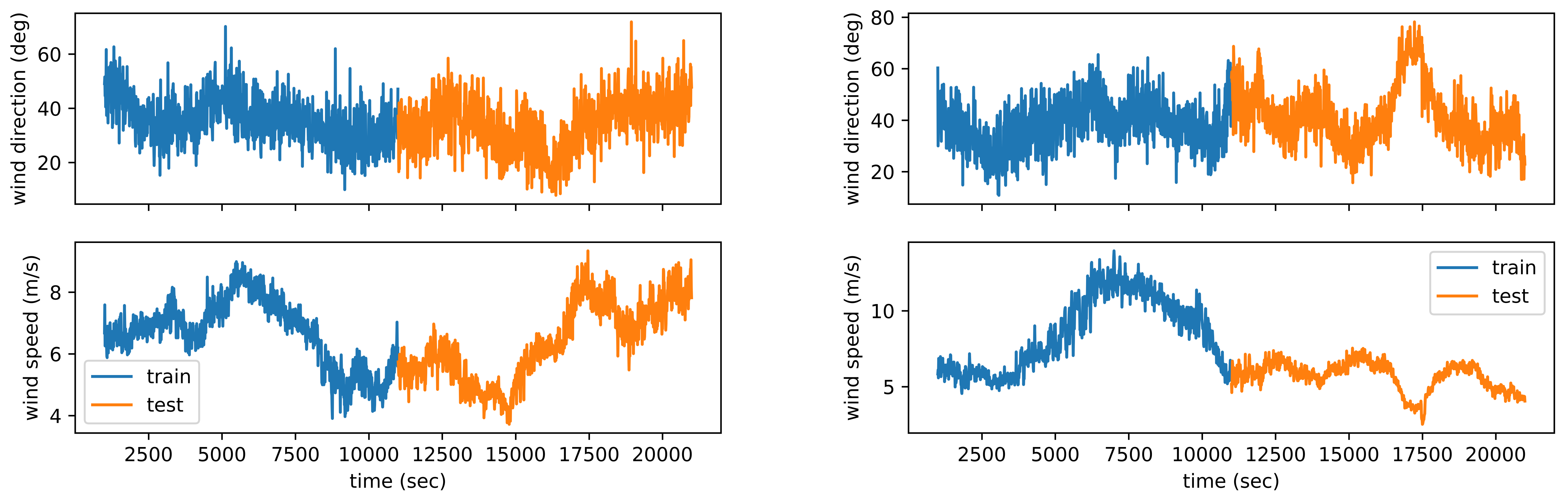}
\caption{The first dataset (left) shows a steady wind direction while the second one (right) shows 2 large wind direction changes. Both are used for our experiments. For each dataset, we train \RLYCA on its first half of a dataset and test it on the other.}
\label{dataset}
\end{figure}
\subsection{Working principle of the yaw system simulation environment }
\begin{figure}[ht]
\centering
\includegraphics[width=\textwidth]{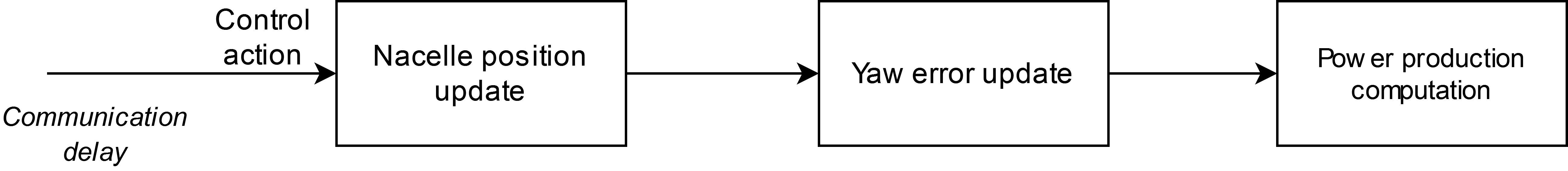}
\caption{Working principle of the yaw system environment}
\label{sim_diagram}
\end{figure}

The working principle of the yaw system environment used for the RL agent is described in Fig.~\ref{sim_diagram}. We update the new nacelle position $\theta_{t+1}$ as in \ref{increment}, where $r$ denotes the yaw rate and $p$ the duration of the control cycle.
\begin{equation}
\label{increment}
\theta_{t+1} = \theta_{t} + p (action-1) r
\end{equation}
The new yaw misalignment is then computed using the new nacelle position and wind direction. We compute the power assuming perfect yaw alignment using the power curve shown in Fig.~\ref{power_curve} and the power under the effect of yaw misalignment is obtained from Eq.~\eqref{loss}, provided that $P_*$ belongs to region 2 of the power curve. Otherwise, we assume no power loss ($P_\gamma = P_*$). We simulate a communication delay of a duration equal to the period of the control cycle. We can assume that the yaw mechanism receives the control signal $a_t$ when the next control cycle ($t+1$) is executed. In practice, this means that we iterate in the wind time series before we compute the new yaw misalignment.
This greatly contributes to the complexity of the task: the wind conditions can change drastically between the real-time where a given action is decided and its application. 

\section{Experiments and Discussions}
In this section, we compare our novel algorithm \RLYCA with benchmarks \CYCAL and \CYCAS on the two datasets 
which we develop into a simulated environment using open AI gym \cite{openai}. 
\subsection{Implementation and parameters}
 \RLYCA uses a PPO agent, implemented in PyTorch \cite{pytorch} and stablebaselines3 \cite{stable-baselines3}. We use an Actor-Critic policy with a 
 2 dense layers of 64 neurons for the value network and 2 layers of 64 neurons for the policy network; trained over 
 200,000 steps. 
For \CYCAS we obtain the control sequence from a control script developed at DEIF. 
 For \CYCAL 
 we reconstruct its control sequence from the real-world turbine logs that saved the position of the nacelle at the same time as the wind condition used for the experiment was saved.
 
The power curves used for the yaw system simulated environment is shown in Fig.~\ref{power_curve}. Table \ref{parameters} gives the parameters used for the yaw system simulated environment. 
\begin{table}
\centering
\caption{Parameters used for the experiment on both datasets. They are identical to those of the turbine from which the controller logs were obtained (REpower MM82).}
\label{parameters}
\resizebox{0.5\textwidth}{!}{
\begin{tabular}{|l|l|}
\hline
parameter name & value\\
\hline
\multicolumn{2}{|l|}{\textit{environment parameters}} \\
\hline
$k$ number of consecutive stationary action to get reward & 2 \\
$j$ number of lagged values in state & 12\\
$w$ weight of $R^{(2)}$ & 40 \\
$v_r$ rated speed & 14m/s\\
$p$ cycle period & 10s\\
$comm$ communication delay & 10s\\
$P_{yaw drive}$ & 18 kWh \\
\hline
\multicolumn{2}{|l|}{\textit{RL agent parameters}} \\
\hline
learning rate & 0.003\\
number of steps before update & 2048\\
batch size & 64\\
number of epochs & 10\\
discount factor & 0.99\\
generalised Advantage Estimator factor & 0.95\\
\hline
\end{tabular}}
\end{table}
The yaw alignment performances of the algorithms are compared using the average yaw misalignment computed as stated in Eq.~\eqref{yaw_mis}. We assess the power extraction performance using the estimate of the power produced given by the yaw system environment and stated in the previous subsection. Since the three experiments are conducted in the same wind conditions given by the testing dataset, the only variable that we use in the computation of the power output and that differs between the three approaches is the nacelle position, which is the direct result of the applications of the three algorithms. We compare the yaw usage of all methods using three metrics: The proportion of the time spent yawing over the simulated duration of the experiment, the angle covered by the yaw mechanism, and the number of yaw actuations. Additionally, we compute the difference in yaw mechanism power consumption $\Delta(t)$ as shown in Eq.~\eqref{yaw_conso}, where $\delta_{\phi_{RL}}(t)$ and $ \delta_{\phi_{simu}}(t)$ denote the change in nacelle position angle during the control cycle $t$ obtained with \RLYCA and with \CYCAS, respectively. Moreover, $r$ denotes the yaw rate and $P_{yawdrive}$ denotes the power consumption of the yaw mechanism.
\begin{equation}
\label{yaw_conso}\Delta(t) = \frac{\delta_{\phi_{RL}}(t) - \delta_{\phi_{simu}}(t)}{r} P_{yawdrive}
\end{equation}

\subsection{Experiment in steady wind direction conditions (dataset/simulation 1)}

\begin{figure}[!ht]
\centering
\includegraphics[width=0.7\textwidth]{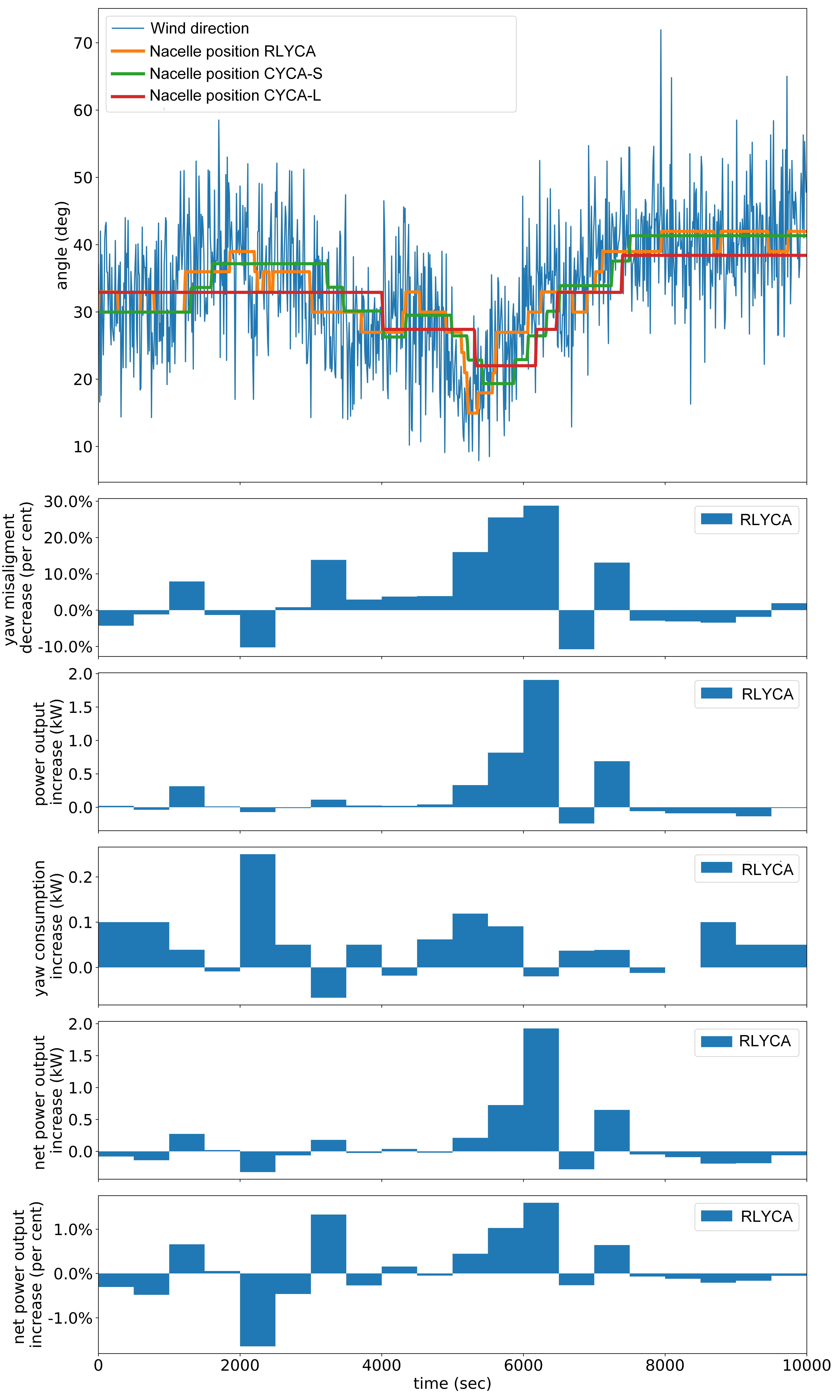}
\caption{Comparison of control signals of the three algorithms on the steady wind direction dataset. The power and yaw error comparisons shown on the bar plots are made against \CYCAS.}
\label{steady_results}
\end{figure}

We first compare the algorithms in steady wind direction conditions. Fig.~\ref{steady_results} shows the nacelle position obtained for the three algorithms. Most of the yaw misalignment decrease is achieved between 5000 and 7000 seconds (first bar plot from top). This results in a large power output increase in the same time interval with a cumulative gain of 1.5kw between 6000 and 6500 seconds. This power production improvement requires more yaw actuations, which translates into a yaw consumption increase that is mainly concentrated in the segment where the wind direction is decreasing.
The yaw usage is also increased in the first half of the dataset but it does not improve the power output. This phenomenon is something that should be corrected in a future version of \RLYCA. The power output change, net of any increase in the yaw consumption, is negative (power loss) in the first half of the dataset because the increase in yaw usage does not lead to a power production increase. Indeed, the power output is increased, but the yaw consumption change is larger than this gain, which results in a negative net power change. The region where the net power output is positive corresponds to a high speed (and thus high power) region so that the relative energy gain is translated into large energy gains. On the opposite, the region where the net power change is negative is a low-speed region, so that the absolute power loss does not affect too much the overall performance of the algorithm. Table \ref{comp_table} shows the results obtained with the three algorithms, while Table \ref{perf_table} summarises the results achieved by \RLYCA. All in all, the algorithm succeeded in allocating more resources to the high-speed segment, resulting in a large power output increase. The relative energy gain achieved by \RLYCA compared to \CYCAS is 0.4 \%, and the net energy gain amounts to 0.31 \%. This gain is due to a better allocation of the yaw resources.
\subsection{Experiment in variable wind direction conditions (dataset/simulation 2)}

\begin{figure}[!ht]
\centering
\includegraphics[width=0.7\textwidth]{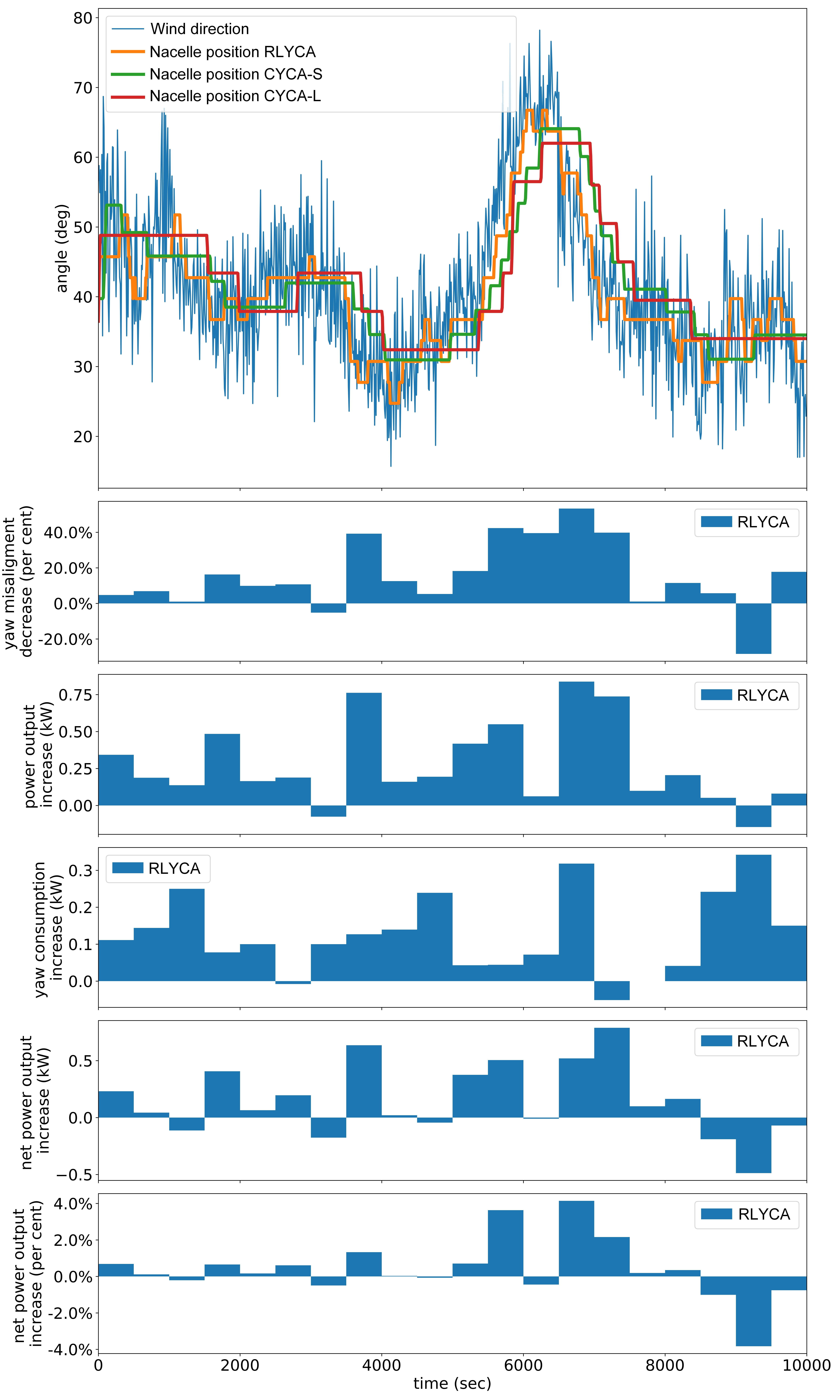}
\caption{Comparison of control signals of the three algorithms on the variable wind direction dataset. The power and yaw error comparisons are made against \CYCAS.}
\label{variable}
\end{figure}

We now compare the algorithms in the simulation derived on the variable wind direction dataset. Fig.~\ref{variable} shows the nacelle position obtained for the three algorithms. Most of the yaw misalignment decrease is achieved in the second half of the dataset.  \RLYCA reacts quicker to the change in wind direction than \CYCAS. 
The power output is increased in almost every region of the dataset. The yaw consumption is also more important but the increase in power output compensates for such additional power consumption everywhere except after 9000 seconds. Indeed, the additional yaw actuations after 9000 seconds do not translate into a better yaw alignment. All in all, the relative energy gain achieved by \RLYCA is 0.75 \%, but the net energy gain is 0.33, as the yaw usage is doubled. The wind speed is stable on the testing set, as shown in Fig.~\ref{dataset}, so that, this time, the energy gain can not be explained by a better allocation of the yaw resources. In this example, it is the overall yaw misalignment decrease achieved by \RLYCA (18.5 \% compared to \CYCAS and 24.4 \% compared to \CYCAL) that explains such a high energy gain.

\begin{table}
\centering
	\caption{Result summary of the three algorithms. Only the power outputs for \CYCAS and \RLYCA are shown as the power output for \CYCAL is impacted by external factors that are not considered in the simulation (e.g. dirt or snow).}
\label{comp_table}
\resizebox{0.7\textwidth}{!}{
\begin{tabular}{|l||l|l|l||l|l|l|}
\hline
&\multicolumn{3}{|l||}{\textit{Steady wind}} & \multicolumn{3}{|l|}{\textit{variable wind}}\\
\hline
& \RLYCA & \CYCAS & \CYCAL & \RLYCA & \CYCAS & \CYCAL \\
\hline
average yaw error (deg) & \textbf{6.18} & 6.52 & 6.91 & \textbf{5.71} & 7.01 & 7.56\\
power output (kWh) & \textbf{1173.1} & 1168.5 & & \textbf{741.5} & 736.0 & \\
angle covered (deg) & 111.0 & 53.5& \textbf{27.3} & 261 & 112.1 & \textbf{91.8} \\
yaw count & 34 & 15 & \textbf{5} & 73 & 27 & \textbf{14} \\
time spent yawing (\%) & 3.7 & 2.0 & \textbf{0.4} & 8.7 & 4.2 & \textbf{1.6}\\
\hline
\end{tabular}}
\end{table}

\begin{table}

\caption{Performance of \RLYCA compared to \CYCAS. The values indicated between parenthesis indicate the comparison with \CYCAL.}
\label{perf_table}
\centering
\resizebox{0.7\textwidth}{!}{
\begin{tabular}{|l||l|l|l||l|l|l|}
\hline
& steady wind & variable wind \\
\hline
average yaw error decrease (\%) & 5.5 (11.2) & 18.5 (24.4) \\
energy gain (\%) & 0.40 & 0.75 \\
net energy gain (\%) & 0.31 & 0.33 \\
\hline
\end{tabular}}
\end{table}

\FloatBarrier

\section{Conclusion and Future Work}
In this paper, we proposed a reinforcement learning yaw control strategy to optimally allocate yaw resources and track wind direction change, with the goal to maximise the power extracted, while keeping yaw usage under a 10 \% bound. We compared our algorithm to a simulation of the current algorithm running on a REpower MM82 and to the control signals gathered from the turbine logs. The comparison was done on two 10000 seconds-long test datasets obtained from past wind measurements logged by the turbine controller. The RL algorithm achieved better wind direction change tracking, which led to a decrease in the yaw misalignment. It also better allocates yaw resources, prioritising high power (and thus high potential gain) segments. Compared to the simulated baseline yaw control algorithm, our new algorithm improves by 5.5 and 18.5 \% the yaw alignment in the steady wind and variable wind conditions, respectively. When comparing its performance to the control sequence obtained from the turbine logs, the last metrics become 11.2 and 24.4 \%. Taking into account the increase in yaw drive consumption, the net energy gain observed on the two datasets was 0.31 and 0.33 \%. For a single 2MW wind class 2 turbine, this amounts to an annual 1500-2500 euros gain\footnote{82 meters rotor diameter, 8-11 GWh annual power production, 80 \% of the time spent in region 2, cosine-cube power loss law, 0.09 euros per kWh.}, which, at the scale of a wind park, represents substantial gains. Throughout our work, we not only showed the relevance of data-driven methods for wind turbines control, but we more generally demonstrated the potential of reinforcement learning to deal with the uncertain nature of renewable energy resources and renewable energy structure behaviour. 
In the future, we plan to use wind turbine digital twins to have an even more accurate simulation of the yaw mechanism, a model of the constraint loads, as well as a more accurate estimation of the power output. The power output could then be directly used in the reward function. We also plan to include data from upstream turbines to the state space, which could help in anticipating the wind changes. Finally, the next step would be to move to the wind farm level, as it has been shown that local yaw alignment can be detrimental at the farm level and because new intended yaw misalignment strategies are being discussed.\\

\noindent{\bfseries Reproducible Paper. }The code and the data sets used for this paper are available at {https://github.com/albanpuech/RL-yaw-control-algorithm-for-wind-turbines}\\

\noindent{\bfseries Acknowledgements. } We would like to thank Mohamed Alami Chehbourne from LIX as well as Walter Telsnig, ‪Anatoliy Zabrovskiy and Martin G{\"o}ldner from DEIF for the helpful discussions and insights on the topic during the preparation of this paper. 

%
%

\bibliographystyle{splncs04}
\bibliography{shortbib.bib}
\end{document}